\documentclass[a4paper,fleqn]{cas-dc}

\usepackage{array,multirow,booktabs}
\usepackage{amsmath}

\usepackage{svg}
\usepackage{amssymb}
\usepackage[authoryear,longnamesfirst]{natbib}

\def\tsc#1{\csdef{#1}{\textsc{\lowercase{#1}}\xspace}}
\tsc{WGM}
\tsc{QE}

\begin{document}
\let\WriteBookmarks\relax
\def\floatpagepagefraction{1}
\def\textpagefraction{.001}

\shorttitle{Forecasting Lithium-Ion Battery Longevity} 

\shortauthors{Hilal et al.}  


\title [mode = title]{Forecasting Lithium-Ion Battery Longevity with Limited Data Availability: Benchmarking Different Machine Learning Algorithms}  



%

\author[]{Hudson Hilal$^*$}[]
\author[1]{Pramit Saha$^*$}[]

\affiliation[1]{organization={Department of Engineering Science, University of Oxford}, city={Oxford},country={United Kingdom}}














\begin{abstract}
As the use of Lithium-ion batteries continues to grow, it becomes increasingly important to be able to predict their remaining useful life. This work aims to compare the relative performance of different machine learning algorithms, both traditional machine learning and deep learning, in order to determine the best-performing algorithms for battery cycle life prediction based on minimal data. We investigated 14 different machine learning models, including Decision Tree, Gradient Boost, Random Forest, K-Nearest Neighbors, AdaBoost, XGBoost, Support Vector Machine, Multi-layer Perceptron, etc. The traditional machine learning algorithms were fed handcrafted features based on statistical data and split into 3 feature groups for testing (Variance, Discharge, and Full).  For deep learning models, we tested a variety of neural network models including different configurations of standard Recurrent Neural Networks (RNN), Gated Recurrent Units (GRU), and Long Short Term Memory (LSTM) with and without attention mechanism. Deep learning models were fed multivariate time series signals based on the raw data for each battery across the first 100 cycles. Our experiments revealed that the machine learning algorithms on handcrafted features performed particularly well, resulting in $10-20\%$ average mean absolute percentage error. The best-performing algorithm was the Random Forest Regressor, which gave a minimum $9.8\%$ mean absolute percentage error. Traditional machine learning models such as Random Forest Regressor excelled due to their capability to comprehend general data set trends. In comparison, deep learning models were observed to perform particularly poorly on raw, limited data. Algorithms like GRU and RNNs that focused on capturing medium-range data dependencies were less adept at recognizing the gradual, slow trends (discharge capacity, temperature) critical for this task.  Our investigation reveals that implementing machine learning models with hand-crafted features proves to be more effective than advanced deep learning models for predicting the remaining useful Lithium-ion battery life with limited data availability.
\end{abstract}


\maketitle
\section{Introduction}
\def\thefootnote{*}\footnotetext{These authors contributed equally to this work}\def\thefootnote{\arabic{footnote}}
\paragraph{}
Lithium-ion (LI) batteries play a large role in powering the modern world. In growing industries today, be it medical equipment or electric vehicles, LI batteries such as Lithium Iron Phosphate (LiFePO4) are a predominant and reliable energy resource. Since their inception, LI batteries have continued to improve in energy density, making them ideal candidates for commercial use \cite{nayak2018lithium}. They are also favored because of their fast charging and longevity. However, one downside to LI batteries is that they experience degradation over time and charging becomes less effective as the battery is continuously used \cite{li2019data}. This degradation is a result of chemical erosion which causes Lithium ions to attach less to the electrode within the battery,. Figure 1 provides a graph demonstrating the degradation of several LI batteries. This image highlights the decreasing discharge capacity graphed against increasing charge cycles.

\begin{figure*}[t]

    \centering
\includegraphics[width=1\columnwidth]{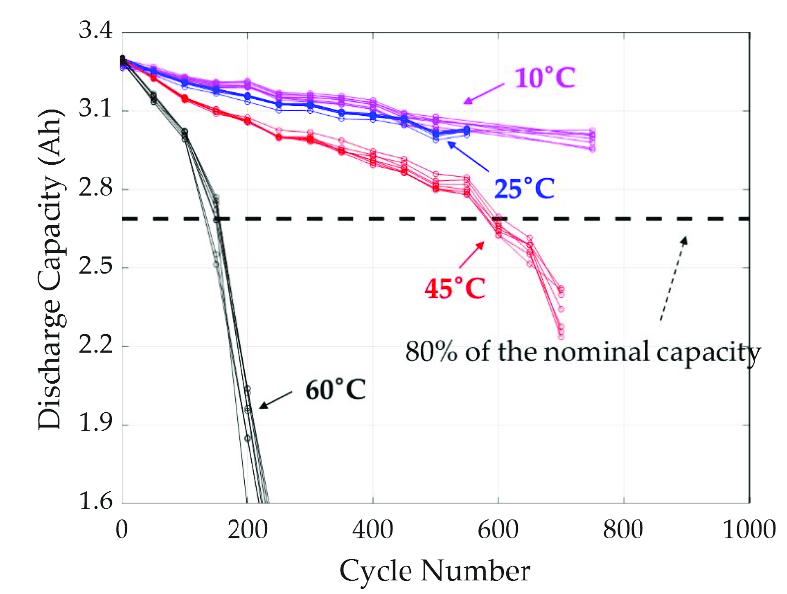}
    \caption{Graph of Battery Cycle Life vs Discharge Capacity \cite{diao2019algorithm}}
\label{fig:1}
\end{figure*}

In order to match the growing demand for renewable energy resources, rechargeable Lithium-ion batteries must be improved upon \cite{etacheri2011challenges}.  Some methods of improvement include examining battery structure, chemistry, and or materials that may affect its longevity and fast-charging speed \cite{liu2019challenges}. One key aspect of battery optimization is the ability to predict a battery’s cycle life, which is useful for tasks such as predictive charging and making production more efficient \cite{muenzel2015multi}. Prediction of LI cycle life opens up many possibilities in battery innovation, benefiting the environment and the economy\cite{hsu2022deep}. However, characteristics such as the nonlinear decay of Lithium-ion batteries makes the prediction of remaining useful cycle life a difficult task  \cite{schuster2015nonlinear}. 

Previous studies have taken different approaches for the task of predicting the remaining cycle life of LI batteries. These approaches fall into two main categories: empirical models and data-driven prediction. For example, in 2008, Safari et al. \cite{safari2008multimodal} used a multi-modal empirical model in order to predict the cycle life of a Lithium-ion battery. This particular study used a solid electrolyte interface in order to model Lithium-ion aging reactions. The study, although successful, displays a fundamental issue with chemical-based models, as they only represent certain battery cell conditions that may not be consistent with other LI batteries. Data-driven models, on the other hand, take a wide range of samples and information to develop accurate prediction based on simple data inputs.

Data-driven prediction of battery cycle life has gained more popularity as machine learning and deep learning algorithms become more advanced. The first evolution of data-based studies utilized recursive filters. For example, Chang et al.'s 2017 study \cite{chang2017new}  created a hybridized unscented Kalman filter that obtained an estimated result that was corrected through a complete ensemble empirical mode decomposition and relevance vector.

More recently, novel computer learning technology has created opportunities for optimizing batteries including material research, chemical simulation, and energy storage. Using machine learning (ML) techniques, data-driven cycle life prediction has been improved significantly  \cite{9714323}. Several ML based studies have been able to predict remaining useful life with high accuracy. One family of promising ML algorithms is artificial neural networks, which leverage the power of deep learning (DL).  Just as data-driven prediction methods shifted from recursive filters to statistical machine learning models, there is now a steady shift toward the use of advanced deep learning models.

Despite the widespread success of data-driven prediction using ML algorithms, accuracy must continue to be increased through rigorous testing and analysis. One caveat is that the training of many ML models requires huge quantities of time and data \cite{burns2013predicting}; by discovering which model architectures and data inputs give the best accuracy, we can improve both speed and accuracy of cycle life prediction as a whole.  Therefore, we worked to find and report the relative success of different machine learning models with regard to accuracy in cycle life prediction.
 
\section{Background}

There are a plethora of works that have used ML for the task of cycle life prediction. Most notable is Severson et al. \cite{severson2019data}, which showed the ability of ML algorithms to predictively model battery life. This study also provided a free-to-use database that recorded the cycle life of 124 LiFePO4 batteries under different charging conditions. This database also serves as a valuable resource for countless research opportunities. In 2019, Shen et al. \cite{shen2019deep} followed in the footsteps of Severson, applying the novelty of deep learning to the problem. This involved the introduction of a convolutional neural network (CNN), a newer, more complex machine learning model. More recently, in 2023, Wan et al \cite{wan2023multi} showed the ability of a convolutional transformer to predict battery temperature by using a time series model. This study highlights the effectiveness of a transformer model in the prediction of a certain battery characteristic during its cycle life. 

Neural networks excel in situations with raw data prediction. Raw data collected from observing battery cycle life can be represented as a multivariate time series, in which there are numerous measurable variables that change with time. This includes temperature, voltage, current, discharge, charge, and many other measurable quantities. As time passes, these variables change responsively and can be measured as such. The time series-like nature of the problem which is battery cycle life prediction allows us to approach the task with new insight. 

\section{Material and Methods}

The ultimate goal of this study was to rigorously test and compare the relative accuracy of machine learning regression and deep learning models for the prediction of LI cycle life.

\subsection{Data Set}

Our data set was based on the information collected and published along with Severson et al. 2019  \cite{severson2019data}. The data set contains extensive information about different statistical and empirical features for 124 LiFePO4  batteries that were tested from maximum capacity until they reached 80 percent of the initial charge/discharge capacity. It is considered to be a benchmark data set regarding the prediction and analysis of battery life. It includes data such as temperature, internal resistance, discharge capacity, charge capacity, and several other measurable data points. This data set was chosen for its size, accessibility, and applicability.

Following in the footsteps of previous cycle life prediction studies, we only used the first 100 cycles for model input. The target data was based on the cycle number at which each battery reached 80\% of its initial discharge capacity. By selecting a smaller number of data points as input, we were truly able to ascertain which methods are most useful for the task of battery life prediction in a real-world setting. Our hope was that he conclusions from this paper are applicable to the real world by using smaller input.

\subsection{Machine Learning Regression Algorithms}

Using the data set of 124 LiFePO4 batteries, we first extracted the relevant features to input in the machine learning models. For each battery cell, we computed various statistical properties based on battery data from the first 100 cycles, all of which are listed below:

\begin{itemize}

    \item Feature 1: Minimum Discharge Difference - Describes the minimum value of the difference between the discharge capacities at cycles 100 and 10.
    \item Feature 2: Variance Discharge Difference - Represents the variance of the difference in discharge capacities between cycles 100 and 10.
    \item Feature 3: Skewness Discharge Difference - Describes the skewness of the discharge capacity change between cycles 100 and 10.
    \item Feature 4:  Kurtosis Discharge Difference  - Captures the kurtosis of the discharge capacity change between cycles 100 and 10.
    \item Feature 5:  Linear Fit Slope - Represents the slope of the linear fit to the capacity fade curve from cycles 2 to 100.
    \item Feature 6: Linear Intercept - Indicates the intercept of the linear fit to the capacity fade curve from cycles 2 to 100.
    \item Feature 7: Discharge Capacity - Represents the discharge capacity at cycle 10.
    \item Feature 8: Maximum Difference -  Represents the difference between the maximum discharge capacity and that at cycle 2.
    \item Feature 9: Average Charge Time -  Represents the average charge time between cycles 2 and 6.
    \item Feature 10: Minimum Resistance - Represents the minimum internal resistance between cycles 2 and 
    \item Feature 11: Resistance Difference - Difference in internal resistance from cycle 10 to 100
\end{itemize}

We then grouped the features listed into different categories for testing to accomplish two goals: 1) to give a range of approaches (in the event of a prediction failure) and 2) to ascertain the effectiveness of different features with ML models. There were three feature groups that we used for input; these were labeled 'Full', 'Discharge', and 'Variance'. Features in each group were chosen based on similar value measurements. For example, Discharge includes all features relating to charge and discharge capacity over several cycles. Each group contains the following handcrafted features:
\begin{itemize}
    \item Full: Feature 1, Feature 2, Feature 5, Feature 6, Feature 7, Feature 9, Feature 10, Feature 11
    \item Discharge: Feature 1, Feature 2, Feature 3, Feature 4, Feature 7, Feature 8
    \item Variance: Feature 2
\end{itemize}

In order to develop a complete understanding of the most effective algorithm using the given data, we selected 14 well-known ML algorithms. The machine learning algorithms tested are listed below.

\begin{enumerate}
    \item \textbf{Linear Regression}: A simple linear model that assumes a linear relationship between input features and the target variable. It estimates coefficients (weights) for each feature to minimize the sum of squared errors between predicted and actual values.
    \item \textbf{Elastic Net}: Combines L1 (Lasso) and L2 (Ridge) regularization techniques. It uses a linear regression model with a combination of L1 and L2 penalty terms.
    \item \textbf{Lasso Regression (Lasso)}: A linear regression model with L1 regularization. Encourages sparsity by adding the absolute values of coefficients as a penalty term to the loss function.
    \item \textbf{Stochastic Gradient Descent (SGD)}: An optimization algorithm is used for training various machine learning models. It updates model parameters iteratively by considering one training example (or a                small batch) at a time.
    \item \textbf{Ridge Regression (Ridge)}: A linear regression model with L2 regularization. Adds the sum of squared coefficients as a penalty term to the loss function.
    \item \textbf{Decision Tree}: A non-linear model that makes decisions by recursively partitioning the feature space.             It divides the data into leaf nodes where predictions are made based on majority class or average target values.
    \item \textbf{Gradient Boost}: An ensemble learning method that combines the predictions of multiple decision trees.           It builds trees sequentially, with each tree correcting the errors of the previous ones.
    \item \textbf{K-Nearest Neighbors (KNN)}: A non-parametric algorithm that classifies data points based on the majority class among their k nearest neighbors in feature space.
    \item \textbf{Random Forest}: An ensemble learning method that builds multiple decision trees and combines their predictions. Each tree is trained on a bootstrap sample of the data and features.
    \item \textbf{AdaBoost}: An ensemble method that combines the predictions of weak learners (often decision trees).             It adjusts the weights of data points to focus on examples that were incorrectly classified by previous models.
    \item \textbf{XGBoost}: An optimized gradient boosting algorithm that is highly efficient and effective for regression and classification tasks.
    \item \textbf{Support Vector Machine (SVM)}: A powerful algorithm for classification and regression tasks.                              It Constructs a hyperplane or set of hyper-planes in a high-dimensional feature space to maximize the margin between classes.
    \item \textbf{RANSAC (Random Sample Consensus)}: An iterative method for robustly estimating parameters of a                 mathematical model from a data set containing outliers.
    \item \textbf{MLP (Multi-layer Perceptron)}: A feed-forward artificial neural network consisting of multiple layers of nodes, designed to learn complex patterns in data by processing information through interconnected neurons.
\end{enumerate}

\subsection{Deep Learning Models}

In this work, we leveraged 3 deep-learning models that are commonly used for sequential data prediction: 

\begin{enumerate}

\item \textbf{Recurrent Neural Network (RNN) }: RNN serves as the foundational form of recurrent neural networks.            It processes sequences by maintaining hidden states and propagating them through sequential time steps. RNN, however, faces challenges related to vanishing gradients, making it less effective for handling lengthy sequences. For more information about the model architecture, readers are referred to a paper exploring the mathematical definition of the RNN, by Sherstinsky et al (2020) \cite{Sherstinsky_2020}.

    \item \textbf{Long Short-Term Memory (LSTM)}: LSTM is a type of recurrent neural network (RNN) designed to address the vanishing gradient problem. It employs a memory cell to store and manage information across long sequences. For more information the readers are referred to the paper defining the LSTM architecture, Hochreiter et al (1997)  \cite{hochreiter1997long}.
    
    \item \textbf{Gated Recurrent Unit (GRU)}: GRU represents a variant of RNN designed to mitigate the vanishing gradient issue. It simplifies the architecture compared to LSTM by combining the input and Forget gates 
        into a single update gate. For more information, the readers are referred to the paper defining GRU architecture, by Chung et al (2014) \cite{chung2014empirical}.

\end{enumerate}

\subsection{Attention-based Mechanisms}

The final approach that we implemented was an attention-based model. Attention-based architecture -- seen in Transformers \cite{vaswani2017attention} -- has become hugely successful in recent years, specifically in fields such as natural language processing and computer vision. Attention mechanisms work through the mapping of query and key-value pairs (all vectors) to an output, from which a weighted sum is calculated and assigned to values in the query and key pairs via a compatibility function. 

We implemented the LSTM, RNN, and GRU models with an attention mechanism in order to investigate their ability to filter data noise effectively or to highlight important patterns within the cycle life. Attention mechanisms are particularly applicable for temporal anomaly detection while also having previous success with other time series models \cite{shih2019temporal}. As such, we ran tests with similar procedures as the aforementioned deep learning models.

We implemented the attention mechanism while defining the LSTM model by defining a linear layer to get attention weights, calculated using the Softmax function. After passing the input data through the initial LSTM layer, we used attention weights to create a context vector, which was calculated by taking the weighted sum of the LSTM output at different time steps. Finally, this context vector is passed through a fully connected layer to get predictions.

\section{Experimental Results}

\subsection{Implementation details}

Each of the aforementioned machine learning algorithms was trained and run through extensive training and testing procedures. Throughout this process, we adjusted model characteristics to get the obtain the highest accuracy possible. After testing and gathering error data regarding the machine learning models, we then shifted our focus to a deep learning approach. 

The feature extraction for deep learning models differed from statistical feature extraction. For these models , we reorganized the features in a time series format, which is typical for prediction and sequential data tasks. This meant that we had to extract the information regarding certain battery features for each cycle that was measured. The target values were given as the cycle number in which the battery reached 80\% of its original discharge capacity. We tested the models on two different feature organizations, one was simply the raw data of discharge values over each cycle, and the other was a multivariate time series with several features including temperature and internal resistance. With the data in collected, we began the testing of our models. 

The training and testing process matched the standard procedure for deep learning models. In order to analyze each model's performance, we adjusted hyperparameters and number of epochs for training several times. The loss function used for backpropagation was mean squared error (MSE). We also recorded the mean absolute percentage error and R-squared over each epoch. We utilized Python’s PyTorch library, which contains a huge number of functionalities useful for our task.

\subsection{Machine Learning Results}

 After preparing the models and completing thorough tests, we collected and examined the prediction data. The main metric we used to quantify a model's success was its mean absolute error percentage (MAPE). Table 1 gives a complete view of the different error margins for each feature grouping (based on mean absolute percentage error).
 
 The first traditional machine learning algorithm tested was the Linear Regression algorithm. In general, it performed sub-optimally, with error margins exceeding 19\% for all feature groups. Next, we tested Elastic Net, Lasso Net, and Ridge regressors. These models are relatively similar and are all linear-based. Both Elastic Net and Lasso Net achieved errors within a range of 13.5 to 11 percent for all feature groupings (which approximately matches the accuracy of Severson et al. testing \cite{severson2019data}). However, their more advanced cousin, the Ridge Regressor, performed poorly, with a mean absolute percentage of 18 or above for each feature grouping. 
 
 The other linear-based machine learning models that were tested included Stochastic Gradient Descent (SGD) and Random Sample Consensus (RANSAC).  The SGD regression algorithm performed poorly, with error margins ranging from 14.3\% to 21\%.  The RANSAC Regressor, on the other hand, was not as promising as the other linear models tested; for the Variance features it had a mean absolute percentage error exceeding 40. This could be because the data set was relatively small and RANSAC relies on a subset of inlier data points, where in this case there are very few. 

Next, we ran the training loop for the rest of the reference machine learning algorithms. The first model among these and the best performing one to be tested was Random Forest. As an ensemble method of decision trees, random forest regression averages the results of a group of decision trees. This model achieved a minimum mean absolute percentage error of about 9.8. However, upon testing this model's simpler cousin, the Decision Tree regressor, we received less accuracy. This model uses a constant approximation of decision rules that it infers from the data. However, rather than being a web of corresponding trees like Random Forest, it applies to the data set on a smaller scale. In testing, the Decision Tree regressor received a minimum error percentage of 13.2, meaning it did not do as well as the Random Forest. 

AdaBoost and Gradient Boosting regressors achieved similar results, with Gradient Boosting regressors outperforming AdaBoost on average. Gradient Boosting regressor performed with a low of 12.3\% error on the Variance features, while AdaBoost achieved a low of 12.8\% on the same input features. For the other feature groupings, however, AdaBoost was significantly worse and received levels of error that exceeded 16\%.

 K-Nearest Neighbors (KNN) regressor and Support Vectors Machine (SVM) models did not render accurate results.  While the SVM regressor received a minimum of 11.8 mean absolute percentage error on the Variance feature grouping, its highest average error was on the Full feature grouping with 34.9\%. On average, the KNN regressor performed better, but its minimum mean accuracy was higher: 18.1\% error on the Variance feature grouping.
 
 Unlike the SVM model, which is less stable, XGBoost performed more consistently on the different  handcrafted features. It performed even better than its similar cousin, the Gradient Boosted regressor, receiving a mean absolute error percentage between 11\% and 19\% on different features. XGBoost is a decision-tree-based regressor, which helps us get a better understanding of why it performed better than other regression models that employ gradient-boosting.

The multi-layer perceptron proved to be  ineffective in terms of prediction based on the handcrafted features. It received high margins of error, especially on the Full feature grouping where it achieved a mean absolute percentage error of 94.6. For the Discharge and Variance feature groupings, the MLP received results of 32.36 and 13.6 mean absolute percentage error, respectively. This poor performance can be attributed to the neural network structure, which was unable to leverage the nuanced relationships between the hand-crafted features.

\begin{center}
\begin{table*}[!ht]
    \centering
    \label{table:nonlin}{
    }
    \caption{Performance comparison of different machine learning regression models in terms of Mean Absolute Percentage Error}
    \begin{tabular}{llll} \hline 
    
        \textbf{Model} &\textbf{ Full}& \textbf{Discharge}& \textbf{Variance}\\ \hline  
        Linear & 22.87& 19.45& 19.5
\\ \hline  
        Elastic Net & 12.43& 11.1 & 12.2
\\ \hline 
        Lasso Net& 13.44& 11.32& 12.3
\\ \hline 
        Ridge & 24.33 & 19.27 & 17.9
\\ \hline  
        SGD& 14.36& 16.51& 21.0\\ \hline  
        Decision Tree& 13.20& 16.38& 19.7 \\ \hline  
        Gradient Boost & 13.19& 12.80& 12.3\\ \hline  
        KNN & 28.63& 22.07& 18.1
\\ \hline  
        Random Forest & 11.13& 10.70& 9.82\\ \hline  
        AdaBoost & 16.01& 14.22& 12.8 \\ \hline  
        XGBoost & 11.32& 14.96& 18.9
\\ \hline  
        SVM & 34.85 & 11.12& 11.8\\ \hline  
        RANSAC & 11.51& 21.80& 41.3\\ \hline  
        MLP& 94.76& 32.62&13.9\\ \hline 
    \end{tabular}

    \label{fig 2}
\end{table*}
\end{center}
In order to verify the accuracy of machine learning models and avoid statistical error, we used two more error metrics in addition to MAPE: R-squared and Mean Squared Error.  These error metrics were tested on each model for multiple runs. This was done to account for the randomness that exists within several of the traditional machine learning architectures and to establish a robust foundation for our analysis of the different models. Additionally, we wanted to provide a complete view of how each model really performs with respect to the data. Table 2 provides the values for each different error metric on which the regression models were tested. The handcrafted features used for the results shown below came from the aforementioned Discharge group.

\begin{center}
\begin{table*}[!ht]
    \centering
    \label{table:nonlin}{
   
    }
    \caption{Regression performance evaluation in terms of three different metrics}
    \begin{tabular}{llll} \hline 
    
       \textbf{Model} & \textbf{MSE}& \textbf{R-Squared}& \textbf{MAPE} \\ \hline  
        Linear & 98700& -0.667&  19.5\\ \hline  
        Elastic Net &  29500& 0.637&  12.2\\ \hline 
        Lasso Net& 29500& 0.637&  12.3\\ \hline 
 Ridge & 90700& -0.115&  17.9\\ \hline  
 SGD& 189600 ± 86320& 1.33 ± 0.0619&  21.0 ± 3.33\\ \hline  
        Decision Tree& 66400 ± 1630& 0.183 ± 0.201&  19.7 ± 1.92\\ \hline  
        Gradient Boost & 30600 ± 672& 0.623 ± 0.008&  12.3± 0.339\\ \hline  
        KNN & 71500 & 0.120&  18.1\\ \hline  
        Random Forest & 2720 ± 898& 0.664 ± 0.0111&  9.82 ± 0.289\\ \hline  
        AdaBoost & 34100 ± 6450& 0.580 ± 0.0793&  12.8 ± 1.59\\ \hline  
        XGBoost & 57100 & 0.297&  18.9\\ \hline  
        SVM & 24900 ± 2820& 0.693 ± 0.0347&  11.8 ± 0.780\\ \hline  
        RANSAC & 11960000 ± 209000& -146 ± 123&  41.3 ± 12.4\\ \hline  
 MLP& 34400 ± 4990& 0.577 ± 0.0614& 13.9 ± 1.61\\ \hline 
    \end{tabular}
       
    \label{fig 3}
\end{table*}

\end{center}

\subsection{Deep Learning Results}

The purpose of this study is to ascertain which machine learning models, if any, can accurately predict battery cycle life enough to be used on a large -- potentially industrial -- scale. The primary deep learning model tested was the Long Short-Term Memory (LSTM) unit. 

LSTM is competent at prediction for a sequential data set regarding time \cite{smagulova2019survey}, and as a result, we applied them to the task of battery life prediction. To optimize LSTM performance, we organized the data set so that it gave us information regarding the discharge data for every cycle of each battery. By training the LSTM on the first 100 cycles of the discharge time series, we hoped that the LSTM model would adjust its weights and be able to predict at which cycle number the battery reached 80 percent of its original discharge capacity. 

In order to minimize loss, we ran a training loop and testing setup for the LSTM. This achieved a final loss of 316.2\% (mean absolute percentage error) over 100 epochs, each epoch using Adam the optimizer for backpropagation. As a result, we found it necessary to extend the training loop to 500 epochs, in hopes of reaching a far smaller margin of error. However, the results were surprising. Rather than learning based on the inputs, the LSTM slowly incremented until it reached an average that was similar to all of the target numbers (cycle where battery reached 80 discharge capacity). The model, while training, reached a minimum mean absolute percentage error of around 320\%, after which it remained at an 'asymptotic' constant. In testing, the model continued to predict inaccurately, giving outputs within a range of two decimal places every time. By this rationale, we can surmise that the data set using the first 100 discharge capacities for each battery had failed to train the model. 

Figure 2 shows the Mean Squared Error (MSE) of the multivariate LSTM model throughout its training process. The graph highlights the model's inability to learn effectively, as it continuously undulates throughout the training process. This indicates that there was no real learning taking place within the network.

\begin{figure*}[t]

    \centering
    \ \label{Figure 2: MSE Loss for LSTM}
\includegraphics[width=1\columnwidth]{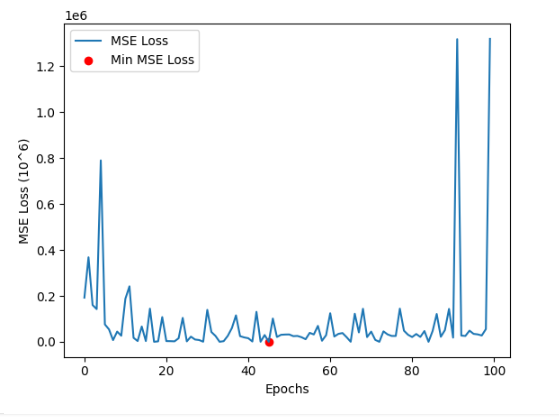}
    \caption{Error curve for multivariate LSTM model throughout the training process indicating failure to converge}
\label{fig 4}
\end{figure*}

The next models tested were the aforementioned Recurrent Neural Network and Gated Recurrent Unit. They performed very similarly to the LSTM in testing. They both received a similar 
'asymptotic' mean absolute percentage error around 300. This indicated an issue with the size or organization of the data set, as none of the diverse model architectures performed with remote success. Even after thorough adjusting of hyperparameters, training time, and loss criterion, the model refused to improve significantly. These tests rendered surprising results, as Recurrent Neural Networks are often the optimal selection for prediction tasks. Results are analyzed further under the Discussion section. To briefly outline, we suspect that it relates to the size of the data or inherent noise within the features.

We also trained a multivariate LSTM on a group of raw features from the data set to attempt another approach to time series prediction regarding deep learning. However, the multivariate approach for the LSTM rendered similar results, which re-emphasized that the issue exists within our data structuring. We chose not to implement a multivariate GRU or RNN without adjusting the features, as they would most likely present similar results.

\begin{center}
\begin{table*}[!ht]
    \centering
    \caption{Performance comparison of different deep learning models in terms of MAPE}
    \label{table:nonlin}{
    }
    \begin{tabular}{llll} \hline 
    
        \textbf{Hidden State Size}& \textbf{LSTM}& \textbf{RNN}& \textbf{GRU} \\ \hline  
        64& 316.23& 316.23& 316.23\\ \hline  
        128& 315.64& 315.64& 315.64\\ \hline 
        256& 313.60& 313.78& 313.65\\ \hline  
 512& 313.64& 311.55& 311.10\\ \hline 
    \end{tabular}
    
    \label{fig 5}
\end{table*}
\end{center}

\subsection{Attention Based Model}

The attention-based LSTM received similar results to its non-attention deep learning counterparts. Attention-based recurrent neural networks typically perform well in time series assignments due to their model architecture and attention function. From our testing results, the Attention LSTM mechanism also failed to learn properly: it received a minimum loss of around 160\% (mean absolute percentage error ). 

This reveals to us that despite performing slightly better -- as attention mechanisms frequently do on sequential data -- deep learning architecture was rendered ineffective for this particular data set. Despite harnessing the power of attention, the deep learning models with raw data were simply unable to learn from the limited scope of data. 

\section{Discussion}

In order to state our claim as to which methodology works best for battery life prediction, we must first examine what factors led to the results visible discrepancies between the performance of statistical regression algorithms and deep learning models. First, it is imperative to discuss the issues that may have been present within the data preparation and organization that negatively impacted testing. The raw data was collected with extreme accuracy and was derived as a part of  Severson et al. 2019's \cite{severson2019data} study. It contains extensive information about the battery characteristics, that are optimal for the training of data-driven models. While the machine learning algorithms performed well at the given task, our deep learning models failed to learn accurately. From these results, we can make multiple hypotheses as to why the DL models lacked accuracy on this data set:
\begin{enumerate}
    \item \textbf{Noise}: Being real-world data. there is a high probability the data includes a large amount of noise, and although it underwent normalization, there is still a chance that this may have impacted results. Noise is often bound to occur in data collection, and this may have particularly impacted our deep learning models, as it makes them prone to errors such as over-fitting. This would have had less of a harsh impact on the machine learning algorithm results, as their error criterion would make it less likely to incorrectly interpret noise.
    \item \textbf{Data Structuring}: While machine learning algorithms often deal with tabular data or statistical values, most deep learning models such as the RNN and LSTM deal with sequential data. This includes speech, text, or other outputs that are sequentially pattern-based. Additionally, the machine learning models chosen are specifically designed for regression tasks, which is also a factor in their success. This indicates that because we chose ML algorithms that are more suitable for the data, they performed better than the DL models that were applied arbitrarily to a regression task.
     \item \textbf{Size}: The most important shortcoming to be considered is the size of the data set. From 124 batteries, our training and testing only used information from the first 100 cycles. For deep learning networks that require more information, this data set is relatively small. While traditional machine learning algorithms also need large amounts of data, the minimum data threshold is far lower than that of neural networks. This may have greatly impacted the results of our study, because the data may have been enough for traditional machine learning algorithms but insufficient for DL models.
\end{enumerate}

Next, it is important to analyze why certain models were successful with this task. For a more comprehensible analysis, we will compare the best-performing models from both the machine learning algorithms and neural networks. From the traditional machine learning approach, one consistently accurate architecture was the Random Forest regressor. The best-performing DL model was the GRU. One of the significant factors in this task that may have led to the success of the Random Forest Algorithm as compared to the GRU is their ability to predict long patterns. While the Random Forest Regressor is keen on recognizing the general trend of a data set, the GRU and RNNs are more effective at capturing medium-range data dependencies. In the context of battery life prediction, the patterns that need to be recognized are slow trends (in terms of discharge capacity, temperature, etc). Therefore, we can see why the regression algorithms may perform better than a recurrent neural network on this task. Memory may have also played a large role in the dominant performance of our machine learning algorithms. Most RNNs, such as a GRU, have a limited memory in comparison to machine learning algorithms. Due to the fact that battery life prediction requires a larger historical context,  Random Forest, for example, may have performed better because it has a larger view of the data.

Finally, we must analyze why certain traditional machine learning models performed better than others.  One noticeable trend is that decision tree-based models achieved higher accuracy for this task. This is observable in the success of the Random Forest regressor and XGBoost regressor. Although Random Forest outperformed the majority of other decision tree models, it is evident that decision tree architecture allowed certain algorithms to battery longevity with precision. This is due in part to their ability to learn without huge amounts of data, which makes them optimal for a task in which only the first 100 cycles are used for regression. Additionally, decision trees are optimal algorithms for handling non-linearity by partitioning the feature space into regions and fitting simple models within each region. Since LI batteries decay non-linearly,  these architectures performed well on this task.

\section{Conclusion}

In this study, we implemented several machine learning algorithms in order to identify the most effective model for the prediction of the life span of LI Batteries. We implemented 14 different machine learning regressors from the skit-learn library including Elastic Net, Lasso Net, SGD, Ridge, Decision Tree, Gradient Boost, KNN, AdaBoost, Random Forest, XGBoost, SVM, RANSAC, and Multi-layer Perceptron. These algorithms proved to be effective on the handcrafted features they were given. We organized the features into 3 distinct statistical categories including Discharge, Full, and Variance features, out of which the Variance category rendered the highest accuracy. There was a distinct pattern that algorithms involving a forest or multi-decision tree architecture typically performed best on the data. Using hand-crafted features and training on the first 100 cycles, we received error margins as low as 9.76 ± 1.1 (mean absolute percentage error) with the Random Forest regressor.

We tested deep learning models specifically known for their ability to process sequential data; the models were fed raw data in the form of univariate and multivariate time series. Each model was trained on a list of time series features including discharge, charge, average temperature, voltage, and current. The models were trained with data from the first 100 cycles of 84 different LiFePO4 batteries. However, deep learning results proved to be less than accurate, rendering extremely high mean absolute percentage error. This was potentially owing to the limited data availability. In other words, 100 cycles for 84 batteries do not provide not enough information to train data-hungry DL models.

From our experimental results, we make several important conclusions: most significantly, the importance of hand-crafted features for the purpose of battery life prediction. This highlights the importance of feature extraction in the process of battery life prediction. The handcrafted statistical features worked well with traditional machine learning algorithms, specifically multi-decision-tree regressors such as Random Forest. The Random Forest regression algorithm's ability to quickly pick up on slow data trends such as decreasing discharge capacity or temperature fluctuations proved effective in this study. Although deep learning is a hugely successful and rapidly developing field, it has its limits. In the context of professional or commercial battery-life prediction, the most optimal prediction method while using minimal data is a machine learning approach. 

Lithium-ion batteries are a crucial energy resource in expanding industries such as electric vehicles and solar energy storage. In order to optimize battery production and usage, the accuracy and data cost of cycle life prediction must be improved. Based on this study, one such possibility for advancement would be the development of deep learning models with fine-tuned architecture fit for battery life prediction. This could be the adaptation of a Temporal Convolutional Network or the use of an active learning loop for selective data modeling. It would also be beneficial to investigate the selection of hand-crafted features that are the most conducive for accurate regression in terms of battery life prediction. Finally, our work underscores the importance of acquiring additional data to improve the accuracy of predicting the cycle life of Lithium-ion batteries using deep learning models.

\bibliographystyle{cas-model2-names}

\bibliography{cas-refs}

\end{document}